%% file: main.tex
\documentclass{article} 
\usepackage{iclr2019_conference,times}
\usepackage[utf8]{inputenc}
\usepackage{graphicx}
\usepackage{subfig}
\usepackage{mathtools}
\usepackage{amsmath}
\usepackage{color}
 \usepackage{caption}
\usepackage{float}
\usepackage{algorithmic,algorithm}
\usepackage{amssymb}
\usepackage{bm}
\usepackage{url}
\usepackage{tikz}
\usepackage{cite}
\usepackage{booktabs}
\usepackage{hyperref}

\newcommand{\bx}{\bm{x}}
\newcommand{\by}{\bm{y}}

\newcommand{\BX}{\mathbf{X}}
\newcommand{\BZ}{\mathbf{Z}}

\input{math_commands.tex}

\usepackage{hyperref}
\usepackage{url}

\title{Bayesian Graph Convolutional Neural Networks using Non-parametric Graph Learning}

\author{\centerline{Soumyasundar Pal, Florence Regol \& Mark Coates} \\
\centerline{Department of Electrical and Computer Engineering,}\\
\centerline{McGill University, Montr\'eal, Qu\'ebec, Canada.}\\
\centerline{\texttt{\{soumyasundar.pal,florence.robert-regol\}@mail.mcgill.ca, mark.coates@mcgill.ca}} 
}

\iclrfinalcopy 
\begin{document}

\maketitle

\begin{abstract}
  Graph convolutional neural networks (GCNN) have been successfully
  applied to many different graph based learning tasks including node
  and graph classification, matrix completion, and learning of node
  embeddings. Despite their impressive performance, the techniques
  have a limited capability to incorporate the uncertainty in the
  underlined graph structure. In order to address this issue, a
  Bayesian GCNN (BGCN) framework was recently proposed. In this
  framework, the observed graph is considered to be a random
  realization from a parametric random graph model and the joint
  Bayesian inference of the graph and GCNN weights is performed. In
  this paper, we propose a non-parametric generative model for graphs
  and incorporate it within the BGCN framework. In addition to the
  observed graph, our approach effectively uses the node features and
  training labels in the posterior inference of graphs and attains
  superior or comparable performance in benchmark node classification
  tasks.
\end{abstract}

\section{Introduction}
\label{sec:intro}
Application of convolutional neural networks to the analysis of data with an
underlying graph structure has been an active area of research in
recent years. Earlier works towards the development of Graph Convolutional Neural Networks
(GCNNs)
include~\citep{bruna2013,henaff2015,duvenaud2015}. ~\citep{defferrard2016}
introduced an approach based on spectral filtering, which was adapted
in subsequent works~\citep{levie2017,chen2018,kipf2017}. On the other
hand, spatial filtering or aggregation strategies are considered
in~\citep{atwood2016,hamilton2017b}. ~\citep{monti2017} present a
general framework for applying neural networks on graphs and
manifolds, which encompasses many existing approaches.

Several modifications have been proposed in the literature to improve the
performance of GCNNs. These include incorporating attention
nodes~\citep{velivckovic2018}, gates~\citep{li2016b,bresson2017}, edge
conditioning and skip
connections~\citep{sukhbaatar2016,simonovsky2017}. Other approaches
consider an ensemble of graphs~\citep{anirudh2017}, multiple adjacency
matrices~\citep{such2017}, the dual graph~\citep{monti2018} and random
perturbation of the graph~\citep{sun2019}. Scalable training for large
networks can be achieved through neighbour
sampling~\citep{hamilton2017b}, performing importance
sampling~\citep{chen2018} or using control variate based stochastic
approximation~\citep{chen2018c}.

Most existing approaches process the graph as if it represents the
true relationship between nodes.  However, in many cases the graphs
employed in applications are themselves derived from noisy data or
inaccurate modelling assumptions. The presence of spurious edges or the absence
of edges between nodes with very strong relationships in these noisy
graphs can affect learning adversely. This can be addressed to some
extent by attention mechanisms ~\citep{velivckovic2018} or generating
an ensemble of multiple graphs by erasing some
edges~\citep{anirudh2017}, but these approaches do not consider
creating any edges that were not present in the observed graph.

In order to account for the uncertainty in the graph structure,
~\citep{zhang2019} present a Bayesian framework where the observed
graph is viewed as a random sample from a collection described by a
parametric random graph model.  This permits joint inference of the
graph and the GCNN weights. This technique significantly outperforms
the state-of-the-art algorithms when only a limited amount of training
labels is available. While the approach is effective, choosing an
appropriate random graph model is very important and the correct
choice can vary greatly for different problems and datasets. Another
significant drawback of the technique is that the posterior inference
of the graph is carried out solely conditioned on the observed
graph. As a result any information provided by the node features and
the training labels is completely disregarded. This can be highly
undesirable in scenarios where the features and labels are highly
correlated with the true graph connectivity.

In this paper, we propose an alternative approach which formulates the
posterior inference of the graph in a non-parametric fashion,
conditioned on the observed graph, features and training
labels. Experimental results show that our approach obtains impressive
performance for the semi-supervised node classification task with a
limited number of training labels.

The rest of the paper is organized as follows. We review the GCNN in
Section~\ref{sec:gcnn} and present the proposed approach in
Section~\ref{sec:methodology}. The numerical experiments are described
and the results are discussed in
Section~\ref{sec:experimental_results}. The concluding remarks are
summarized in Section~\ref{sec:conclusion}.

\section{Graph convolutional neural networks (GCNNs)}
\label{sec:gcnn}

Although graph convolutional neural networks have been applied in a
variety of inference tasks, here we consider the node classification
problem in a graph for conciseness. In this setting, we have access to
an observed graph $\mathcal{G}_{obs} = (\mathcal{V},\mathcal{E})$,
where $\mathcal{V}$ is the set of $N$ nodes and $\mathcal{E}$ denotes
the set of edges. For every node $i$ we observe a feature vector
$\bx_i \in \mathbf{R}^{d \times 1}$, but the label $\by_i$ is known for
only a subset of the nodes $\mathcal{L}\subset \mathcal{V}$. The goal is to
infer the labels of the remaining nodes using the information provided
by the observed graph $\mathcal{G}_{obs}$, the feature matrix
$\BX = [\bx_1, \bx_2, \dots, \bx_N]^T$ and the training labels
$\mathbf{Y_{\mathcal{L}}} = \{\by_i: i \in \mathcal{L}\}$.

A GCNN performs graph convolution operations within a neural network
architecture to address this task.  Although there are many different versions of the graph convolution operation, the layerwise propagation rule for the simpler architectures~\citep{defferrard2016,kipf2017} can be expressed as:
\begin{align}
\mathbf{H}^{(1)} &=\sigma(\mathbf{\hat{A}}_{\mathcal{G}}\mathbf{X}\mathbf{W}^{(0)}) \,,\\
\mathbf{H}^{(l+1)} &= \sigma(\mathbf{\hat{A}}_{\mathcal{G}}\mathbf{H}^{(l)}\mathbf{W}^{(l)})\,. 
\end{align}
Here $\mathbf{H}^{(l)}$ are the output features from layer $l-1$, and
$\sigma$ is a pointwise non-linear activation function. The normalized adjacency operator
$\mathbf{\hat{A}}_{\mathcal{G}}$, which is derived from the observed graph, 
determines the mixing of the output features across the graph at each
layer. $\mathbf{W}^{(l)}$ denotes the weights of the neural network at layer
$l$. We use $\mathbf{W}=\{\mathbf{W}^{{l}}\}_{l=1}^L$ to denote all GCNN weights.

For an $L$-layer network, the final output is $\mathbf{Z} =
\mathbf{H}^{(L)}$. Learning of the weights of the neural network is carried out
by backpropagation with the objective of minimizing an error metric between
the training labels $\mathbf{Y}_{\mathcal{L}}$ and the network predictions $\mathbf{Z}_{\mathcal{L}} = \{z_i : i \in {\mathcal{L}}\}$, at the nodes in the training set.

\section{Methodology}
\label{sec:methodology}
We consider a Bayesian approach, by constructing a joint posterior
distribution of the graph, the weights in the GCNN and the node
labels. Our goal is to compute the marginal posterior probability of
the node labels, which is expressed as follows:
\begin{align}
p(\BZ|\mathbf{Y_{\mathcal{L}}},\BX,\mathcal{G}_{obs}) 
= \int p(\BZ|\mathbf{W},\mathcal{G}_{obs},\BX) p(\mathbf{W}|\mathbf{Y_{\mathcal{L}}},\BX,\mathcal{G}) p(\mathcal{G}|\mathcal{G}_{obs}, \BX, \mathbf{Y_{\mathcal{L}}}) \,d\mathbf{W}\,d\mathcal{G}\label{eq:exact_posterior}\,.
\end{align}
Here $\mathbf{W}$ denotes the random weights of a Bayesian GCNN over
graph $\mathcal{G}$. In a node classification setting, the term
$p(\BZ|\mathbf{Y_{\mathcal{L}}},\BX,\mathcal{G}_{obs})$ is modelled
using a categorical distribution by applying a softmax function to the
output of the GCNN. As the integral in~\eqref{eq:exact_posterior} can not be computed analytically, a Monte Carlo approximation is formed as follows:
\begin{align}
p(\BZ|\mathbf{Y_{\mathcal{L}}},\BX,\mathcal{G}_{obs}) \approx 
\dfrac{1}{S} \sum_{s=1}^S
 \dfrac{1}{N_G}\sum_{i=1}^{N_G} p(\BZ|\mathbf{W}_{s,i},\mathcal{G}_{obs},\BX)\,.
\label{eq:MC_posterior}
\end{align}
In this approximation, $N_G$ graphs $\mathcal{G}_{i}$ are sampled from $p(\mathcal{G}|\mathcal{G}_{obs}, \BX, \mathbf{Y_{\mathcal{L}}})$ and $S$ weight samples $\mathbf{W}_{s,i}$ are drawn from $p(\mathbf{W}|\mathbf{Y_{\mathcal{L}}},\BX,\mathcal{G}_{i})$ by training a Bayesian GCN corresponding to the graph $\mathcal{G}_{i}$.

\citep{zhang2019} assume that $\mathcal{G}_{obs}$
is a sample from a collection of graphs associated with a parametric
random graph model and their approach targets inference of
$p(\mathcal{G}|\mathcal{G}_{obs})$ via marginalization of the random
graph parameters, ignoring any possible dependence of the graph
$\mathcal{G}$ on the features $\BX$ and the labels
$\mathbf{Y_{\mathcal{L}}}$. By contrast, we consider a non-parametric
posterior distribution of the graph $\mathcal{G}$ as
$p(\mathcal{G}|\mathcal{G}_{obs}, \BX, \mathbf{Y_{\mathcal{L}}})$. 
This allows us to incorporate the information provided by the
features $\BX$ and the labels $\mathbf{Y_{\mathcal{L}}}$ in the graph
inference process.

We denote the symmetric adjacency matrix with non-negative entries of
the random undirected graph $\mathcal{G}$ by $A_{\mathcal{G}}$ . The prior distribution for $\mathcal{G}$ is defined as
 \begin{align}
    p(\mathcal{G}) \propto \begin{dcases} \exp{(\alpha\mathbf{1}^T\log(A_{\mathcal{G}}\mathbf{1}) - \beta \|A_{\mathcal{G}} \|_F^2)},\quad \text{if } A_{\mathcal{G}} \geq \mathbf{0}, A_{\mathcal{G}} = A_{\mathcal{G}}^T\,\\
    0, \quad  \text{otherwise}\label{eq:prior}\,.
    \end{dcases}
\end{align}
For allowable graphs, the first term in the log prior prevents any isolated node in
$\mathcal{G}$ and the second encourages low weights for the
links. $\alpha$ and $\beta$ are hyperparameters which control the
scale and sparsity of $A_{\mathcal{G}}$. The joint likelihood of
$\BX$, $\mathbf{Y_{\mathcal{L}}}$ and $\mathcal{G}_{obs}$ conditioned
on $\mathcal{G}$ is:
 \begin{align}
    p(\BX,\mathbf{Y_{\mathcal{L}}},\mathcal{G}_{obs}|\mathcal{G}) \propto \exp{(- \|A_{\mathcal{G}} \circ Z(\BX,\mathbf{Y_{\mathcal{L}}},\mathcal{G}_{obs}) \|_{1,1})}\,,\label{eq:joint_likelihood}
\end{align}
where
$Z(\BX,\mathbf{Y_{\mathcal{L}}},\mathcal{G}_{obs}) \geq \mathbf{0}$ is
a symmetric pairwise distance matrix between the nodes. The symbol
$\circ$ denotes the Hadamard product and $\|\cdot\|_{1,1}$ denotes the
elementwise $\ell_1$ norm. We propose to use
 \begin{align}
  Z(\BX,\mathbf{Y_{\mathcal{L}}},\mathcal{G}_{obs}) = Z_{1} (\BX,\mathcal{G}_{obs}) + \delta Z_{2}(\BX,\mathbf{Y_{\mathcal{L}}},\mathcal{G}_{obs})\,,\label{eq:distance}   
 \end{align}
 where, the $(i,j)$'th entries of $Z_1$ and $Z_2$ are defined as follows: 
 \begin{align}
     Z_{1,ij} (\BX,\mathcal{G}_{obs}) &= \|\ve_i - \ve_j\|^2\,,\label{eq:dist1}\\
     Z_{2,ij}(\BX,\mathbf{Y_{\mathcal{L}}},\mathcal{G}_{obs}) &= \frac{1}{|\mathcal{N}_i||\mathcal{N}_j|}\displaystyle{\sum_{k \in \mathcal{N}_i}\sum_{l \in \mathcal{N}_j}}\mathbf{1}_{(\hat{c}_k \neq \hat{c}_l)}\,.\label{eq:dist2}
 \end{align}
 Here, $\ve_i$ is any suitable embedding of node $i$ and $\hat{c}_i$
 is the label obtained at node $i$ by a base classification algorithm. $Z_1$
 summarizes the pairwise distance in terms of the observed topology
 and features and $Z_2$ encodes the dissimilarity in node labels
 robustly by considering the obtained labels of the neighbours in the
 observed graph. In this paper, we choose the Graph Variational
 Auto-Encoder (GVAE) algorithm~\citep{kipf2016} as the node embedding
 method to obtain the $\ve_i$ vectors and use the GCNN proposed by~\citep{kipf2017} as the base
 classifier to obtain the $\hat{c}_i$ values. The neighbourhood is defined as:
 \begin{align}
 \mathcal{N}_i = \{j | (i,j) \in \mathcal{E}_{\mathcal{G}_{obs}}\} \cup\{i\}\,.\nonumber
 \end{align}
 $\delta$ is a hyperparameter which controls the importance of $Z_2$ relative to $Z_1$. We set:
 \begin{align}
 \delta = \dfrac{\displaystyle{\max_{i,j}} Z_{1,ij}}{\displaystyle{\max_{i,j}} Z_{2,ij}}\,.\nonumber
 \end{align}
 In order to use the approximation in~\eqref{eq:MC_posterior}, we need to obtain samples from the posterior of $\mathcal{G}$. However, the design of a suitable MCMC in this high dimensional
 ($\mathcal{O}(N^2)$, where $N$ is the number of the nodes) space is
 extremely challenging and computationally expensive. Instead we
 replace the integral over $\mathcal{G}$ in~\eqref{eq:exact_posterior} by the maximum a posteriori estimate of $\mathcal{G}$, following the approach of~\citep{mackay1996}. We solve the following optimization
 problem
 \begin{align}
 \hat{\mathcal{G}} &= \argmax_{\mathcal{G}} p(\mathcal{G}|\mathcal{G}_{obs}, \BX, \mathbf{Y_{\mathcal{L}}})\,,\label{opt:graph_inference}
 \end{align}
and approximate the integral in~\eqref{eq:exact_posterior} as follows:
 \begin{align}
     p(\BZ|\mathbf{Y_{\mathcal{L}}},\BX,\mathcal{G}_{obs})  \approx \frac{1}{S}\displaystyle{\sum_{s=1}^S}p(\BZ|\mathbf{W}_s,\mathcal{G}_{obs},\BX)\,.\label{eq:approx_posterior_map}
 \end{align}
 Here, $S$ weight samples $\mathbf{W}_s$ are drawn from $p(\mathbf{W}|\mathbf{Y_{\mathcal{L}}},\BX,\hat{\mathcal{G}})$. The MAP inference in~\ref{opt:graph_inference} is equivalent to learning a $N \times N$ symmetric adjacency matrix of $\hat{\mathcal{G}}$.
 \begin{align}
 A_{\hat{\mathcal{G}}}&= \argmin_{\substack{A_{\mathcal{G}} \in \mathbf{R_+}^{N\times N}, \\ A_{\mathcal{G}}=A_{\mathcal{G}}^T }} \|A_{\mathcal{G}} \circ Z\|_{1,1} -\alpha\mathbf{1}^T\log(A_{\mathcal{G}}\mathbf{1}) + \beta\|A_{\mathcal{G}}\|_F^2 \,.\label{opt:adj_inference}
 \end{align}
 The optimization problem in~\ref{opt:adj_inference} has been studied
 in the context of graph learning from smooth
 signals. In~\citep{kalofolias2016}, a primal-dual optimization
 algorithm is employed to solve this problem. However the complexity
 of this approach scales as $\mathcal{O}(N^2)$, which can be
 prohibitive for large graphs. We use the approximate algorithm in
 ~\citep{kalofolias2017}, which has an approximate 
 $\mathcal{O}(N\log N)$ complexity. This formulation allows us to
 effectively use one hyperparameter instead of $\alpha$ and $\beta$ to
 control the sparsity of the solution and provides a useful heuristic
 for choosing a suitable value.
 
 Various techniques such as expectation propagation~\citep{hernandez2015}, variational inference~\citep{gal2016,sun2017,louizos2017}, and Markov Chain Monte Carlo methods~\citep{neal1993,korattikara2015,li2016d} can be employed for the posterior inference of the GCNN weights. Following the approach in~\citep{zhang2019}, we train a GCNN over the inferred graph $\hat{\mathcal{G}}$ and use Monte Carlo dropout~\citep{gal2016} to sample $\mathbf{W}_s$ from a particular variational approximation of $p(\mathbf{W}|\mathbf{Y_{\mathcal{L}}},\BX,\hat{\mathcal{G}})$. The resulting algorithm is described in Algorithm~\ref{alg:bgcn_non_param}.

 {\color{red}
 
\begin{algorithm}[tb]
\caption{Bayesian GCN using non-parametric graph learning}
\label{alg:bgcn_non_param}
\begin{algorithmic}[1]
\STATE {\bfseries Input:}  $\mathcal{G}_{obs}$, $\BX$, $\mathbf{Y_{\mathcal{L}}}$
\STATE {\bfseries Output:}  $p(\BZ|\mathbf{Y_{\mathcal{L}}},\BX,\mathcal{G}_{obs})$

\STATE Train a node embedding algorithm using $\mathcal{G}_{obs}$ and $\BX$ to obtain $\ve_i$ for $1 \leq i \leq N$. Compute $Z_1$ using~\eqref{eq:dist1}. 

\STATE Train a base classifier using $\mathcal{G}_{obs}$, $\BX$ and  $\mathbf{Y_{\mathcal{L}}}$ to obtain $\hat{c}_i$ for $1 \leq i \leq N$. Compute $Z_2$ using~\eqref{eq:dist2}. 

\STATE Compute $Z$ using~\eqref{eq:distance}.

\STATE  Solve the optimization problem in~\ref{opt:adj_inference} to obtain $A_{\hat{\mathcal{G}}}$ (equivalently, $\hat{\mathcal{G}}$).

\FOR{$s=1$ {\bfseries to} $S$}
\STATE Sample weights $W_{s}$ using MC dropout by training a GCNN over the graph $\hat{\mathcal{G}}$.
\ENDFOR

\STATE Approximate $p(\BZ|\mathbf{Y_{\mathcal{L}}},\BX,\mathcal{G}_{obs})$ using~\eqref{eq:approx_posterior_map}.
\end{algorithmic}
\end{algorithm}
 
}
 
\section{Experimental Results}
\label{sec:experimental_results}
We investigate the performance of the proposed Bayesian
GCNN on three citation datasets~\citep{sen2008}: Cora,
CiteSeer, and Pubmed. In these datasets each node corresponds to a document and the undirected edges are citation links. Each node has a sparse
bag-of-words feature vector associated with it and the node label represents the topic of the document. We address a semi-supervised node classification task where we have access to the labels of a few nodes per class and the goal is to infer labels for the others. We consider three different experimental settings where we have 5, 10 and 20 labels per class in the training set. The partitioning of the data in 20 labels per class case is the same as in~\citep{yang2016} whereas in the other two cases, we construct the training set by including the first 5 or 10 labels from the previous partition. The hyperparameters of GCNN are borrowed from~\citep{kipf2017} and are used for the BGCN algorithms as well.

We compare the proposed BGCN in this paper with ChebyNet~\citep{defferrard2016}, GCNN~\citep{kipf2017}, GAT~\citep{velivckovic2018} and the BGCN in~\citep{zhang2019}. Table~\ref{table:accuracy} shows the summary of results based on 50 runs with random weight initializations. 

\begin{figure}[h]
\minipage{0.008\textwidth}
 \includegraphics[scale=0.5, clip, trim={0.7cm 0  0.7cm 1cm}]
 {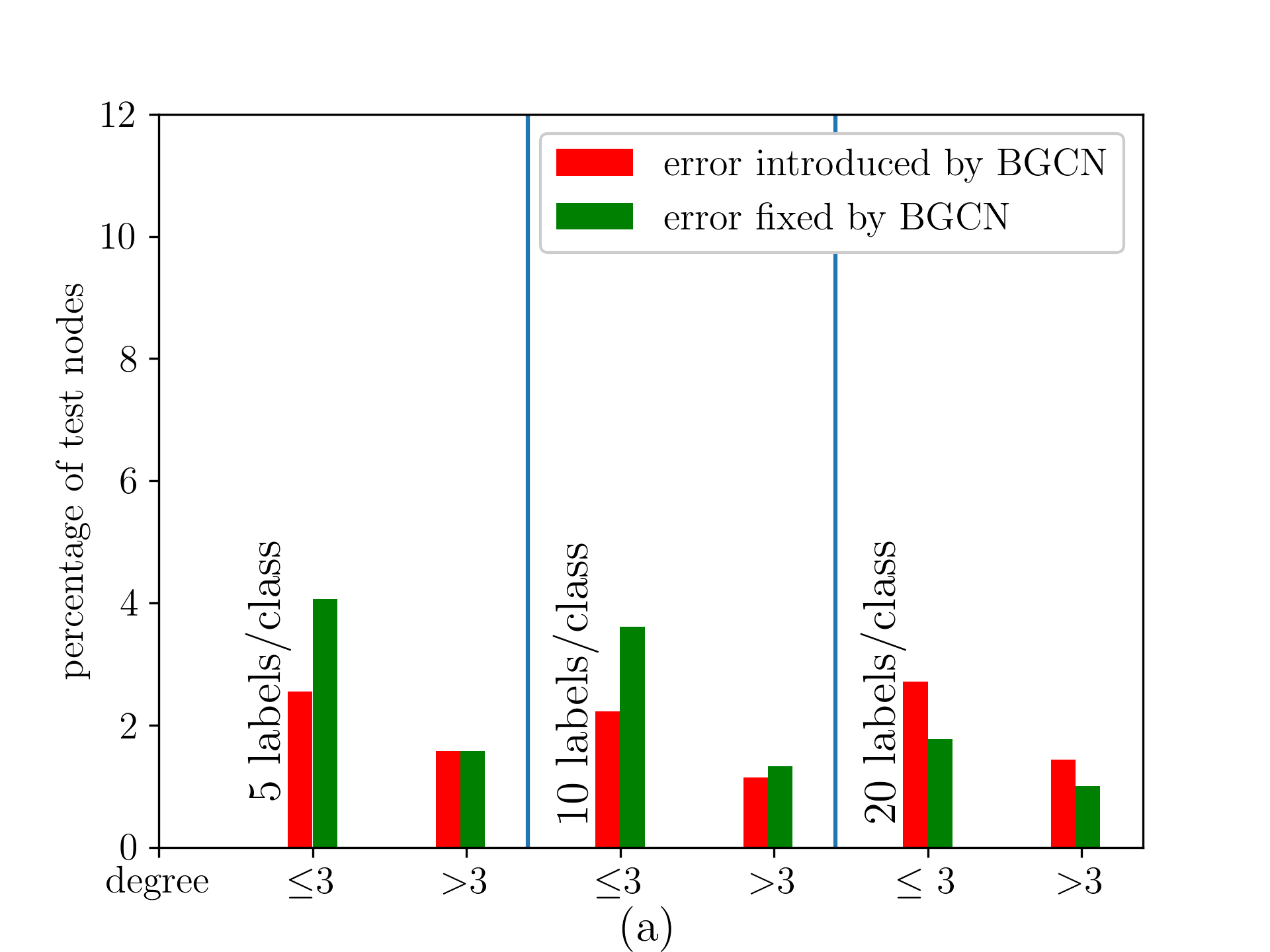}
 \captionsetup{justification=centering}
\endminipage\hspace{6.8cm}
\minipage{0.008\textwidth}
 \includegraphics[scale=0.5, trim={0.7cm 0 0.7cm 1cm}, clip]
 {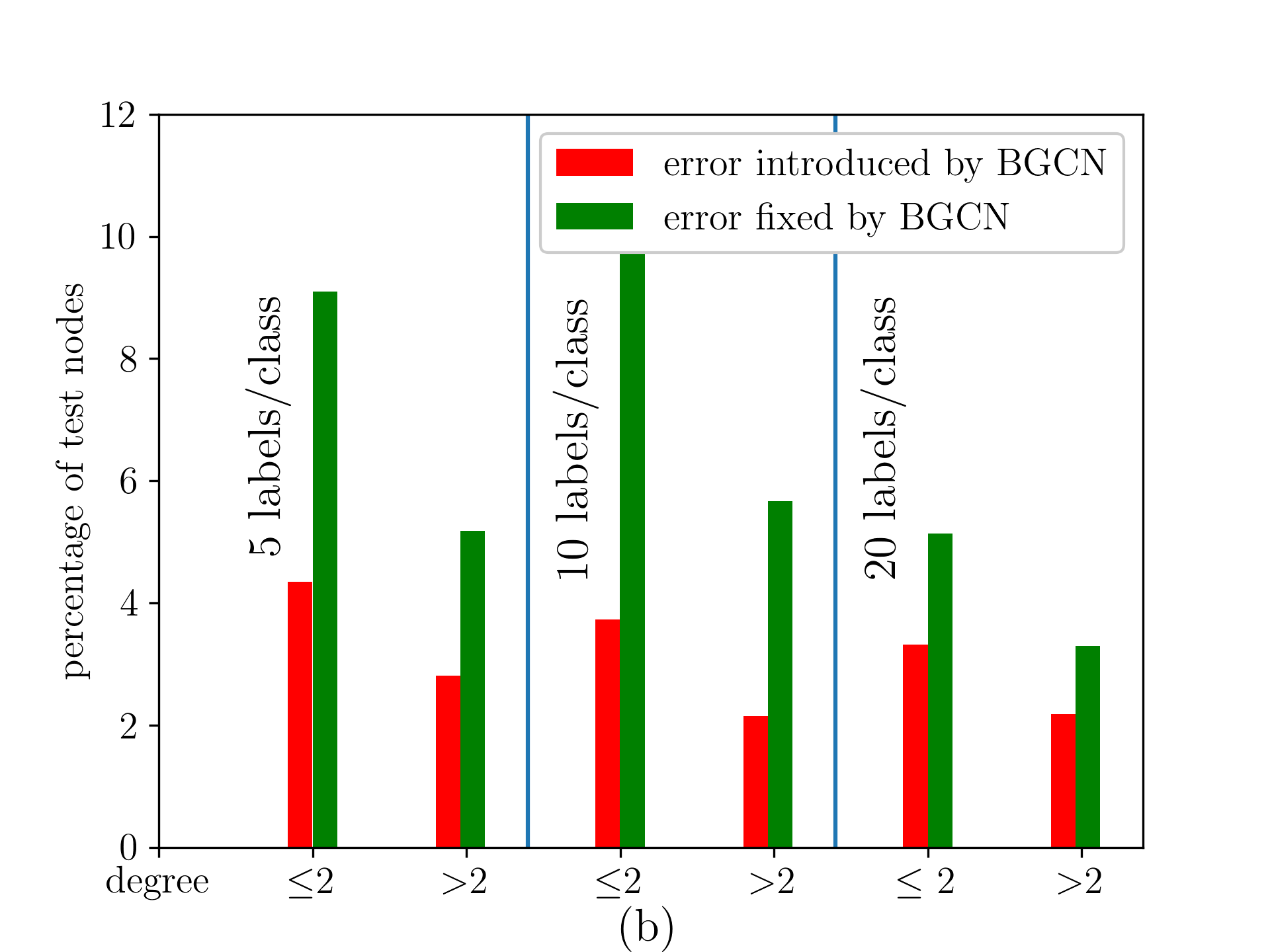}
\endminipage\hfill
  \caption{Barplot of different categories of nodes in the (a) Cora and (b) Citeseer datasets based on the classification results of the GCNN and the proposed BGCN algorithms. The two groups are formed by thresholding the degree of the nodes in the test set at the median value.}
\label{fig:degree_cora}
\end{figure}

\begin{figure}[h]
\minipage{0.008\textwidth}
 \includegraphics[scale=0.5, clip, trim={0.7cm 0  0.7cm 1cm}]
 {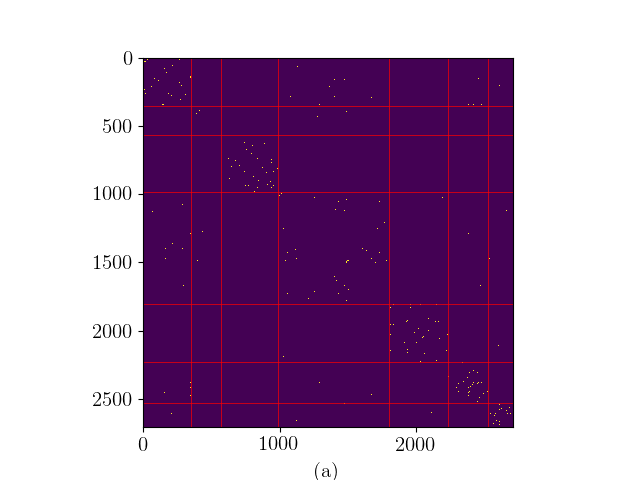}
 \captionsetup{justification=centering}
\endminipage\hspace{6.8cm}
\minipage{0.008\textwidth}
 \includegraphics[scale=0.5, trim={0.7cm 0 0.7cm 1cm}, clip]
 {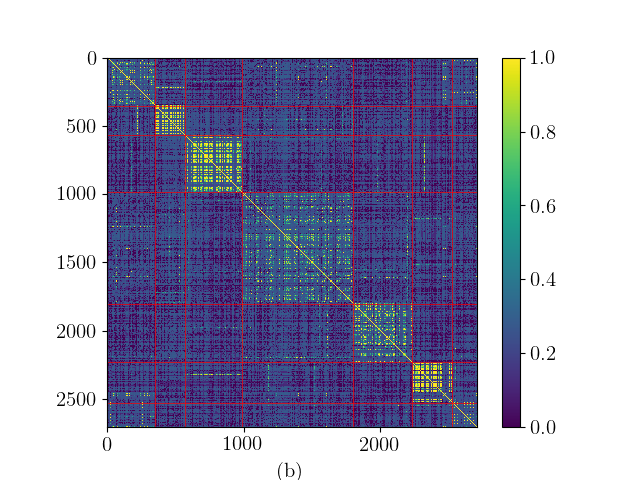}
\endminipage\hfill
  \caption{(a) the observed adjacency matrix ($A_{\mathcal{G}_{obs}})$ and (b) the MAP estimate of adjacency matrix ($A_{\hat{\mathcal{G}}}$) for the Cora dataset. }
\label{fig:adj}
\end{figure}

\begin{table}[ht]
\setlength{\tabcolsep}{2pt}
\small
\begin{tabular}{|c|c|c|c|c|c|c|c|c|c|}
\hline
\textbf{Dataset}      & \multicolumn{3}{c|}{\textbf{Cora}} & \multicolumn{3}{c|}{\textbf{Citeseer}} & \multicolumn{3}{c|}{\textbf{Pubmed}} \\ \hline
\textbf{Labels/class} & \textbf{5}      & \textbf{10}      & \textbf{20}     & \textbf{5}       & \textbf{10}       & \textbf{20}       & \textbf{5}       & \textbf{10}      & \textbf{20}      \\ \hline
\textbf{ChebyNet}             &67.9$\pm$3.1            &72.7$\pm$2.4             &80.4$\pm$0.7         &53.0$\pm$1.9            &67.7$\pm$1.2              &70.2$\pm$0.9        &68.1$\pm$2.5            &69.4$\pm$1.6              &76.0$\pm$1.2             \\
			\textbf{GCNN}             &74.4$\pm$0.8            &74.9$\pm$0.7              &\textbf{81.6$\pm$0.5}       &55.4$\pm$1.1            &65.8$\pm$1.1              &70.8$\pm$0.7      &69.7$\pm$0.5            &72.8$\pm$0.5            &78.9$\pm$0.3  \\
			\textbf{GAT}              &73.5$\pm$2.2            &74.5$\pm$1.3              &81.6$\pm$0.9   &55.4$\pm$2.6            &66.1$\pm$1.7              &70.8$\pm$1.0    &70.0$\pm$0.6   & 71.6$\pm$0.9            &76.9$\pm$0.5 \\
			\textbf{BGCN}     &75.3$\pm$0.8   &76.6$\pm$0.8    &81.2$\pm$0.8   &57.3$\pm$0.8   &70.8$\pm$0.6     &72.2$\pm$0.6 &70.9$\pm$0.8      &72.3$\pm$0.8    &76.6$\pm$0.7\\
			\textbf{BGCN (ours)}     &\textbf{76.0$\pm$1.1}   &\textbf{76.8$\pm$0.9}   &80.3$\pm$0.6   &\textbf{59.0$\pm$1.5}   &\textbf{71.7$\pm$0.8}     &\textbf{72.6$\pm$0.6}  &\textbf{73.3$\pm$0.7}      &\textbf{73.9$\pm$0.9}    &\textbf{79.2$\pm$0.5}\\
			\hline
\end{tabular}
	\caption{Classification accuracy (percentage of correctly predicted labels) for the three datasets.}
 	\label{table:accuracy}
\end{table}
The results in Table~\ref{table:accuracy} show that the proposed algorithm yields higher classification accuracy compared to the other algorithms in most cases. Figure~\ref{fig:degree_cora} demonstrates that in most cases, for the Cora and the Citeseer datasets, the proposed BGCN algorithm corrects more errors of the GCNN base classifier for low degree nodes. The same trend is observed for the Pubmed dataset as well. In Figure~\ref{fig:adj}, the adjacency matrix ($A_{\hat{\mathcal{G}}}$) of the MAP estimate graph $\hat{\mathcal{G}}$ is shown along with the observed adjacency matrix $A_{\mathcal{G}_{obs}}$ for the Cora dataset. We observe that compared to $A_{\mathcal{G}_{obs}}$, $A_{\hat{\mathcal{G}}}$ has denser connectivity among the nodes with the same label.

\vspace{-1em}

\vspace{-0.25em}

\section{Conclusion}
\label{sec:conclusion}
In this paper, we present a Bayesian GCNN using a non-parametric graph inference technique. The proposed algorithm achieves superior performance when the amount of available labels during the training process is limited. Future work will investigate extending the methodology to other graph based learning tasks, incorporating other generative models for graphs and developing scalable techniques to perform effective inference for those models.

\bibliography{references}
\bibliographystyle{iclr2019_conference}

\end{document}

%% file: math_commands.tex
\usepackage{amsmath,amsfonts,bm}

\def\eqref#1{equation~\ref{#1}}

\def\1{\bm{1}}

\def\ve{{\bm{e}}}

\DeclareMathAlphabet{\mathsfit}{\encodingdefault}{\sfdefault}{m}{sl}
\SetMathAlphabet{\mathsfit}{bold}{\encodingdefault}{\sfdefault}{bx}{n}

\DeclareMathOperator*{\argmax}{arg\,max}
\DeclareMathOperator*{\argmin}{arg\,min}